# A Combination of K-Nearest Neighbor and Deep Neural Networks for Credit Card Fraud Detection

**Dinara Rzayeva**
Kapital Bank OJC
Small and Medium Enterprises Department
Baku, Azerbaijan
dinararj@gmail.com

**Saber Malekzadeh**
GI Academy
Data Science Department
Baku, Azerbaijan
smalekzp@gmail.com

*Abstract-* Detection of a Fraud transaction on credit cards became one of the major problems for financial institutions, organizations and companies. As the global financial system is highly connected to non-cash transactions and online operations fraud makers invent more effective ways to access customers' finances. The main problem in credit card fraud detection is that the number of fraud transactions is significantly lower than genuine ones. The aim of the paper is to implement new techniques, which contains of under-sampling algorithms, K-nearest Neighbor Algorithm (KNN) and Deep Neural Network (KNN) on new obtained dataset. The performance evaluation showed that DNN model gives precise high accuracy (98.12%), which shows the good ability of presented method to detect fraudulent transactions.

## I. INTRODUCTION

Modern financial system is connected to the intensive development of non-cash payments. Consumers nowadays have convenient methods to pay for almost all their purchase-using credit cards. From the moment the non-cash payment systems and mechanisms were created, fraud makers regularly try to invent the new ways to access someone's finances illegally. All transactions can easily be completed online by only entering consumers' credit card information. However, in this case consumers are put at risk when a data breach leads to monetary theft. On the other hand, financial institutions and merchants can lose customers' loyalty and company's reputation.

Machine Learning techniques are used to develop computational methodologies, which can detect non-legitimate transactions based on amount and time of those transactions. Fraud detection, for credit cards, is the method of distinguishing between transactions. Two categories of transaction can happen either genuine or fraudulent.
Typical fraud detection systems encompass associate academic degree automatic tool and a manual technique. The automatic tool depends on fraud detection rules. It analyses all the new incoming transactions and assigns a fallacious score. Fraud investigators produce the manual technique. [1]
Essentially the biggest problem in dealing with Credit Card Fraud Detection is that real data is hardly ever available for exploration due to the issue of confidentiality.

Another very important challenge in fraud detection is that on global level fraudulent transactions are amounted to less than 0,03% of the total payment transactions. This ratio is detained in the distribution of any credit card fraud dataset resulting in highly imbalanced classes. [2]

If this problem would not be taken into consideration, any algorithm that classifies correctly genuine transactions would show accuracy level above 99%, disregarding the fact that all the minority class transaction are classified falsely.
Fraudulent behavior tends to alter over time in order to avoid detection. Thus, Credit Card Fraud predictive models should not be static. From point of view of Machine Learning, this problem is known as concept drift.

Fraud detection is also a cost-sensitive problem. It means that the cost of misclassifying genuine transaction is different from the cost of misclassifying fraudulent ones. [2]

## II. RELATED WORK

Imbalanced dataset is a common problem faced in machine learning, since most traditional machine learning classification model can't handle imbalanced dataset. High misclassification cost often happened on minority class, because classification model will try to classify all the data sample to the majority class.

Under-sampling refers to a group of techniques designed to balance the class distribution for a classification dataset that has a skewed class distribution.

An imbalanced class distribution will have one or more classes with relatively few examples (the minority classes) and one or more classes with relatively many examples (the majority classes).

Under-sampling techniques remove examples from the training dataset that belong to the majority class in order to better balance the class distribution, such as reducing the skew from a 1:100 to a 1:10, 1:2, or even a 1:1 class distribution. This is different from oversampling that involves adding examples to the minority class in an effort to reduce the skew in the class distribution.

The main algorithms that can provide the better performance are described below [3]:

*Edited Nearest Neighbors (ENN)*

In ENN, under-sampling of the majority class is done by removing points whose class label differs from a majority of its k nearest neighbors.

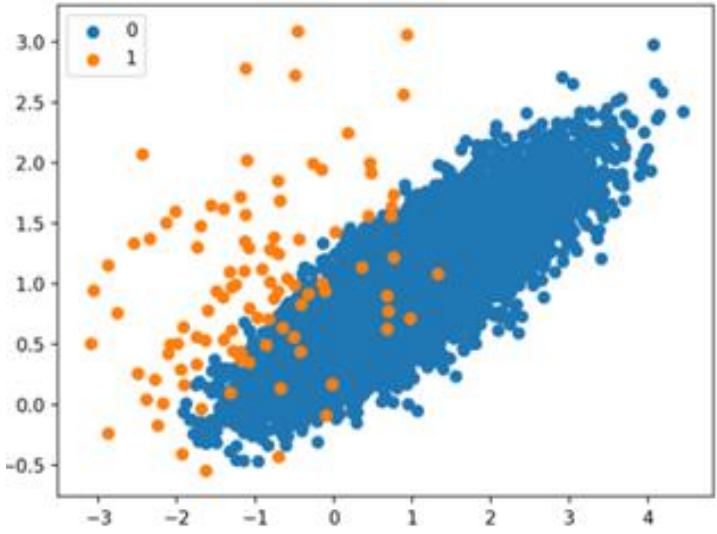

*Figure 1: Scatter Plot of Imbalanced Dataset Under-sampled with ENN*

In Repeated Edited Nearest Neighbor, the ENN algorithm is applied successively until ENN can remove no further points.

One-Sided Selection, or OSS for short, is an under-sampling technique that combines Tomek Links and the Condensed Nearest Neighbor (CNN) Rule.

Specifically, Tomek Links are ambiguous points on the class boundary and are identified and removed in the majority class. The CNN method is then used to remove redundant examples from the majority class that are far from the decision boundary.

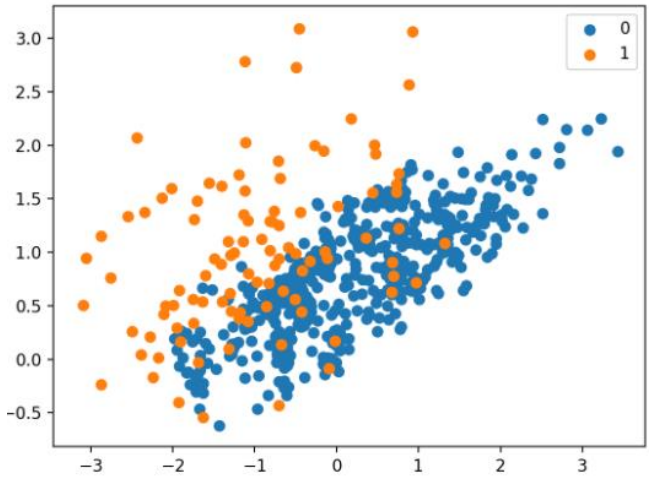

*Figure 2 Scatter Plot of Imbalanced Dataset Under-sampled with One-sided Selection. [3]*

The approach involves first selecting all examples from the minority class. Then all of the ambiguous examples in the majority class are identified using the ENN rule and removed. Finally, a one-step version of CNN is used where those remaining examples in the majority class that are misclassified against the store are removed, but only if the number of examples in the majority class is larger than half the size of the minority class.

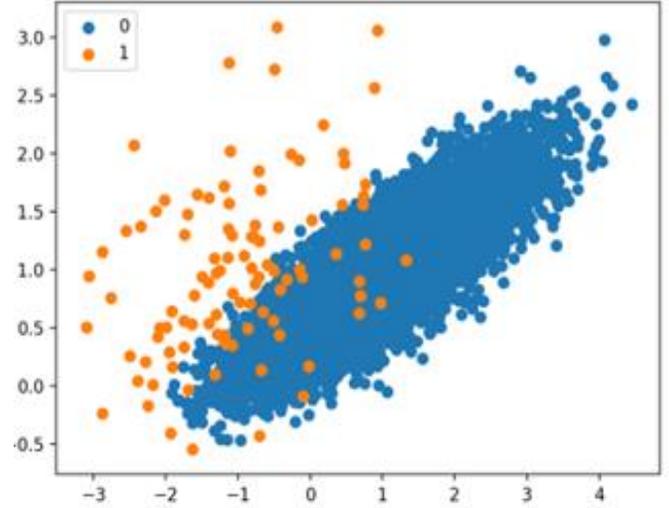

*Figure 3 Scatter Plot of Imbalanced Dataset Under-sampled with NCR. [3]*

K-nearest Neighbor (KNN) works extremely well in credit card fraud detection systems using supervised learning techniques. Outlier detection (OD) is used to detect unusual behavior of a system using a different mechanism. Malini implemented KNN algorithm and outlier detection methods to optimize the best solution for the fraud detection problem; he showed that KNN could suit for detecting fraud with the limitation of memory. [4]

Credit Card Fraud Detection with K-Nearest Neighbor requires a distance to measure between two data instances. It is based on measuring the distances between the test data and each of the training data to decide final classification output. In KNN, any incoming transaction is checked/classified to the corresponding nearest point of the previously held transaction point. Then, if the nearest point is real or fraud, the incoming transaction is judged accordingly. [5]

This method consists of using the k nearest points to the one we aim to predict [2]. A specific norm is used to measure the distance between points. The new observation is assigned the class with the majority of the k nearest points. The norm usually used to measure the distance between two observations p and q (an observation is $\in$ n ) is the Euclidean distance (1):

$$d(x,y) = \sqrt{\sum_{i=1}^{n}(x_i - y_i)^2} \quad (1)$$

In [7] authors have done a comparative analysis on several majorly known and used algorithms namely Naïve Bayes, k-Nearest Neighbor and Logistic Regression. Matthews Correlation Coefficient and Balanced Classification Rate measure the algorithms. Taking into consideration the fact that they are performing on highly imbalanced credit card fraud dataset, it presented sufficient results 96.82%, 96.69% and 54.86% for Naïve Bayes, k-Nearest Neighbor and Logistic Regression respectively, which in this case we can see that k-Nearest Neighbor has the highest accuracy. [7]

An alternative to ML strategies is the use of deep neural networks. To overcome the problem of over- and/or under-sampling, deep neural network can be discussed. It is built with the deep learning library TensorFlow, with the high-level neural networks Keras API on top. TensorFlow is an open-source machine learning library. Keras works on a higher level and is designed to enable fast experimentation. It is user friendly, modular, and easy to extend. It is possible to build deep neural nets with only a few lines of code.

The network consists of an input layer, output layer, and in-between (multiple) hidden layers. Each layer consists of a number of nodes that are connected with adjacent layers; all these connections have weights associated with them. In DNN, the information moves from the input layer to the output layer via the hidden layers in the middle, and there are no loops in the network. The hidden layers compute a series of transformations that change the similarities between cases and ensure that each layer has a non-linear relation to adjacent layers. [7]

## III. METHOLOGY

For the purpose of fraud detection, credit card transactions dataset from Kaggle was used [9]. This dataset contains 284,807 credit card transactions, which were performed by cardholders in September 2013 by European cardholders. Each transaction is represented by: 30 principal components extracted from the original data; the time from the first transaction; the amount of money.

The transactions have two labels: one for fraudulent and zero for genuine transactions. Only 492 transactions in the dataset are fraudulent.

The data contains principal components instead of the original transaction features for confidentiality reasons.

### 3.1 Data preprocessing

While processing the huge dataset, features selection is an important aspect, here selecting the variables, which are most relevant for future analysis. The quality of preprocessing reduces over fitting and training time and improves accuracy. In case of implemented dataset. As most of the features anonymized two features "Time" and Amount" were observed  It was concluded that "Time" feature have the same pattern in both fraudulent and genuine transactions. So it was concerned not import for the purpose of classification. Analysis of 'Amount' future showed that all the transactions that are more 10 000 are genuine. Therefore, this feature can be taken into consideration. Unwanted feature 'Time' is dropped.

On this stage, the data normalization and scale of features were produced.

### 3.2 Data sampling algorithm implementation

As the dataset is highly imbalanced, the under-sampling technics was implemented. Four following algorithms were tested in order to get the data that are more balanced:
- Repeated Edited Nearest Neighbors
- Edited Nearest Neighbors
- One-Sided Selection
- Neighborhood Cleaning Rule

The purpose of abovementioned rules was to find ambiguous and noisy examples in a dataset and reduce the majority class. Obtained reduced dataset was divided into train and test dataset in proportion of 80:20.

### 3.3 K-Nearest Neighbor classifier

The K-Nearest Neighbor classifier (KNN) was implemented for classification. For KNN algorithm, the n-neighbors parameter was set as 5.

### 3.4. Deep Neural Network model architecture

The first deep learning network is a feed-forward network with five hidden layers. In Figure 4, the implemented network is summarized. Two dense layers, dropout layers are implemented to prevent overfitting of the model. The first and second dropout layers possess respectively, dropout values of 0.5 and 0.3.

To supervise the neural network, 100 iterations were performed on the corresponding data set. The batch size was settled as 32.

```
Layer (type)                 Output Shape              Param #   
=================================================================
dense_124 (Dense)            (None, 16)                480       
_________________________________________________________________
dense_125 (Dense)            (None, 24)                408       
_________________________________________________________________
dropout_37 (Dropout)         (None, 24)                0         
_________________________________________________________________
dense_126 (Dense)            (None, 20)                500       
_________________________________________________________________
dense_127 (Dense)            (None, 15)                315       
_________________________________________________________________
dropout_38 (Dropout)         (None, 15)                0         
_________________________________________________________________
dense_128 (Dense)            (None, 24)                384       
_________________________________________________________________
dense_129 (Dense)            (None, 1)                 25        
=================================================================
Total params: 2,112
Trainable params: 2,112
Non-trainable params: 0
```

*Figure 4.DNN Model Structure*

## IV. EVALUATION AND RESULTS

After implementing under-sampling techniques, examples from the majority class were removed consisting of both redundant examples and ambiguous examples. The proportion of the new dataset was 1:10

It can be observed from *Figure 5* that the fraud data points have been moved towards one cluster, whereas there are only few fraud transaction data points are there among the normal transaction data points.

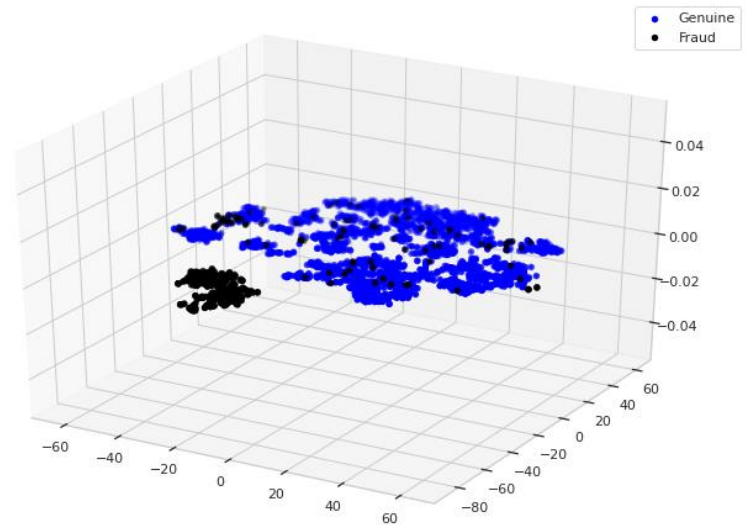

Figure 5: The Dimensionality Plot

Confusion matrix constructed for the models demonstrates that DNN model overcome KNN model in task to identify most of the fraudulent transactions correctly.

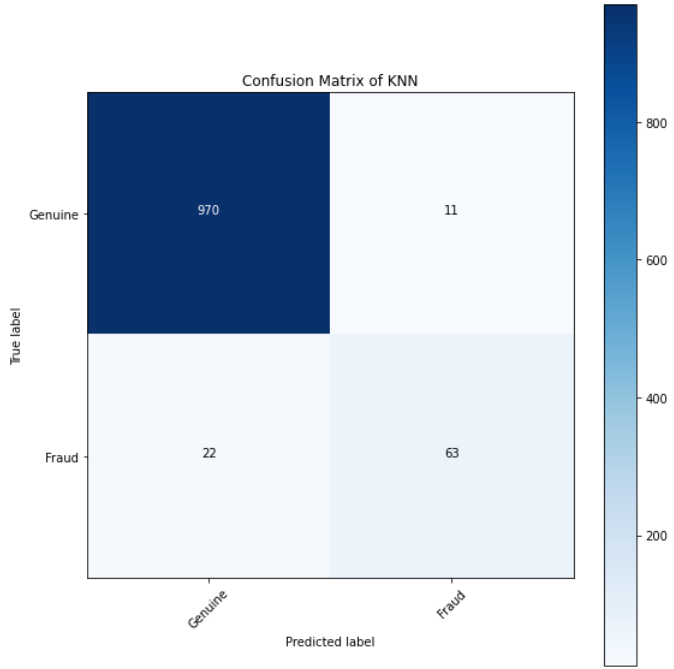

Figure 6: Confusion Matrix for KNN model

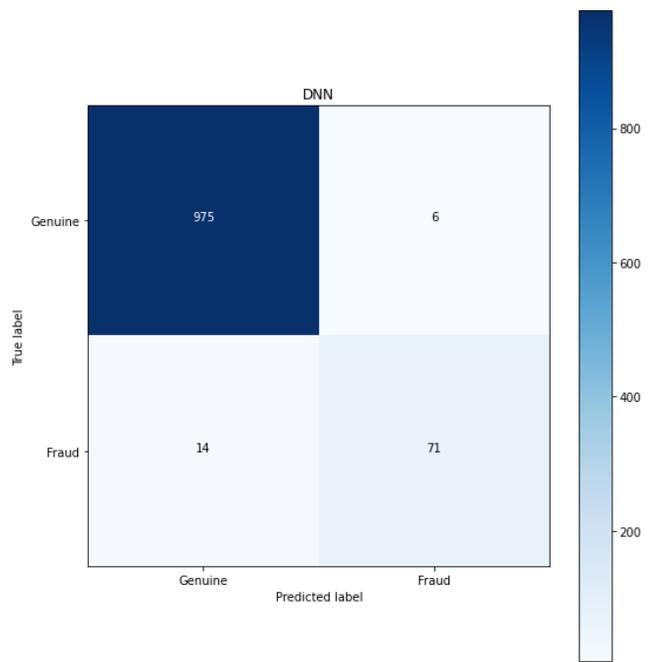

Figure 7: Confusion Matrix for DNN model

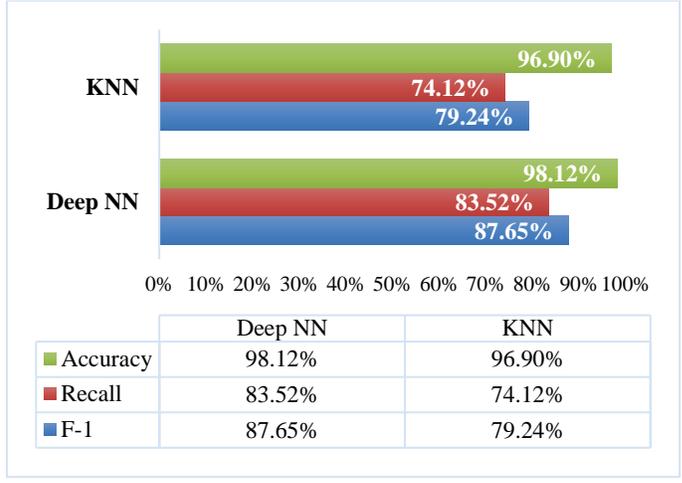

Figure 8: The comparison of results of implemented algorithms

Base on methodology of the research, the Area of under Curve was used to identify the success rate of the model. As the percentage of AUC of both algorithms implemented are high, (91.54% - for KNN model and 94.02% - for DNN model), that means that we found unsupervised learning rate with true positive rate on our model. (Figure 9 and Figure 10). However, implemented DNN model is more perspective in detecting fraud transactions.

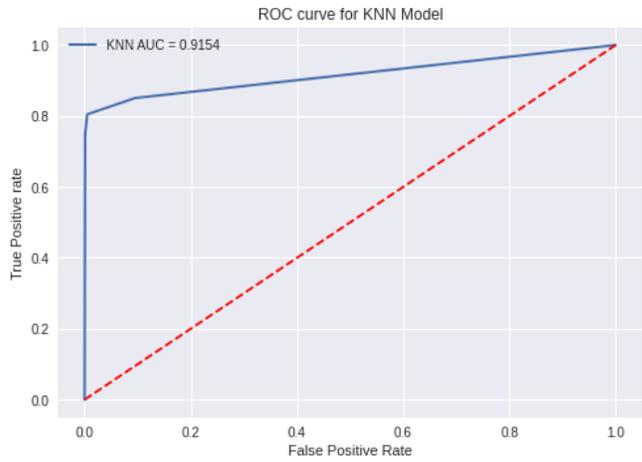

Figure 9: ROC curve of implemented KNN model

In Figure 8, it can be observed that deep neural network demonstrated higher accuracy and recall rate comparing to KNN algorithm.

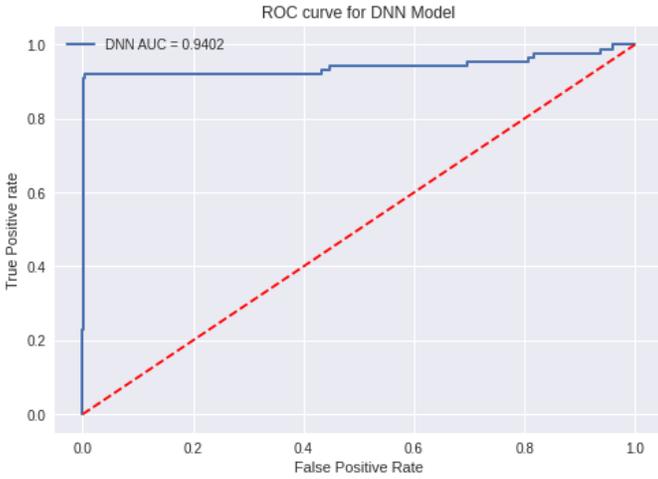

*Figure 10: ROC curve of implemented DNN model*

| Models | Accuracy |
|---|---|
| KNN *[10]* | 97.15% |
| KNN *[7]* | 97.92% |
| KNN+OD *[4]* | 97.25% |
| AE+RBM *[12]* | 97.05% |

*Table 1: Accuracy rates for fraud detection models presented in other research papers*

## V. CONCLUSION

The aim of this research was to propose two new methods that could improve the classification performance for credit card fraud detection. We can conclude that resampling can improve the performance significantly. After experiments it was determined that One-Sided Selection method gives most promising results in this case.

Comparing ML and deep learning methods it is obvious that DNN are the way to explore complex features within the data so that the model can learn better to predict frauds more efficiently with less false alarms.

As future work, it can be determined that another type of deep neural network neural Autoencoder have a great opportunity to build a fraud detector even in the absence of fraudulent transactions. The idea stems from the more general field of anomaly detection and works very well for fraud detection. A neural Autoencoder with more or less complex architecture is trained to reproduce the input vector onto the output layer using only "normal" data — in our case, only genuine transactions.